\pdfoutput=1
\documentclass[conference]{IEEEtran}
\usepackage{times}

\usepackage{graphicx}
\usepackage{amsmath}
\usepackage{amssymb}

\usepackage[numbers]{natbib}
\usepackage{multicol}
\usepackage[bookmarks=true]{hyperref}

\usepackage{soul}
\usepackage{comment}

\usepackage{subcaption}
\usepackage{mwe}
\usepackage{balance}
\usepackage{multirow}

\usepackage{siunitx} 

\renewcommand{\b}[1]{\boldsymbol{#1}} 

\usepackage{placeins} 

\usepackage{xcolor}
\definecolor{guidance_color}{HTML}{4D4D4D}
\definecolor{local_color}{HTML}{344154}
\definecolor{decision_color}{HTML}{990000}

\newcounter{RCcnt}
\newcounter{subRCcnt}

\begin{document}

\title{A Vehicle System for Navigating Among Vulnerable Road Users Including Remote Operation}


\author{\IEEEauthorblockN{O. de Groot, A. Bertipaglia, H. Boekema, V. Jain, M. Kegl, V. Kotian, T. Lentsch, Y. Lin,\\ C. Messiou, E. Schippers, F. Tajdari, S. Wang, Z. Xia, M. Zaffar, R. Ensing,  M. Garzon, J. Alonso-Mora,\\ H. Caesar, L. Ferranti, R. Happee, J.F.P. Kooij,  G. Papaioannou, B. Shyrokau, and D.M. Gavrila}
\IEEEauthorblockA{Department of Cognitive Robotics, Delft University of Technology}}

\maketitle

\begin{abstract}
    We present a vehicle system capable of navigating safely and efficiently around Vulnerable Road Users (VRUs), such as pedestrians and cyclists. The system comprises key modules for environment perception, localization and mapping, motion planning, and control, integrated into a prototype vehicle. A key innovation is a motion planner based on Topology-driven Model Predictive Control (T-MPC). The guidance layer generates multiple trajectories in parallel, each representing a distinct strategy for obstacle avoidance or non-passing. The underlying trajectory optimization constrains the joint probability of collision with VRUs under generic uncertainties.
    To address extraordinary situations ("edge cases") that go beyond the autonomous capabilities — such as construction zones or encounters with emergency responders — the system includes an option for remote human operation, supported by visual and haptic guidance.
    
    In simulation, our motion planner outperforms three baseline approaches in terms of safety and efficiency. We also demonstrate the full system in prototype vehicle tests on a closed track, both in autonomous and remotely operated modes.   
\end{abstract}

\IEEEpeerreviewmaketitle

\section{Introduction}
\label{sec:introduction}

Automated driving has made steady progress in recent years. For instance, advanced highway autopilot systems now enable drivers to divert their attention and engage in side activities—until prompted to retake control (i.e., conditional automation). At the same time, driverless vehicles (``robotaxis'') have begun offering mobility services within geo-fenced areas in several U.S. cities (high automation).

\begin{figure}[h]
    \centering
    \includegraphics[width=0.9 \columnwidth]{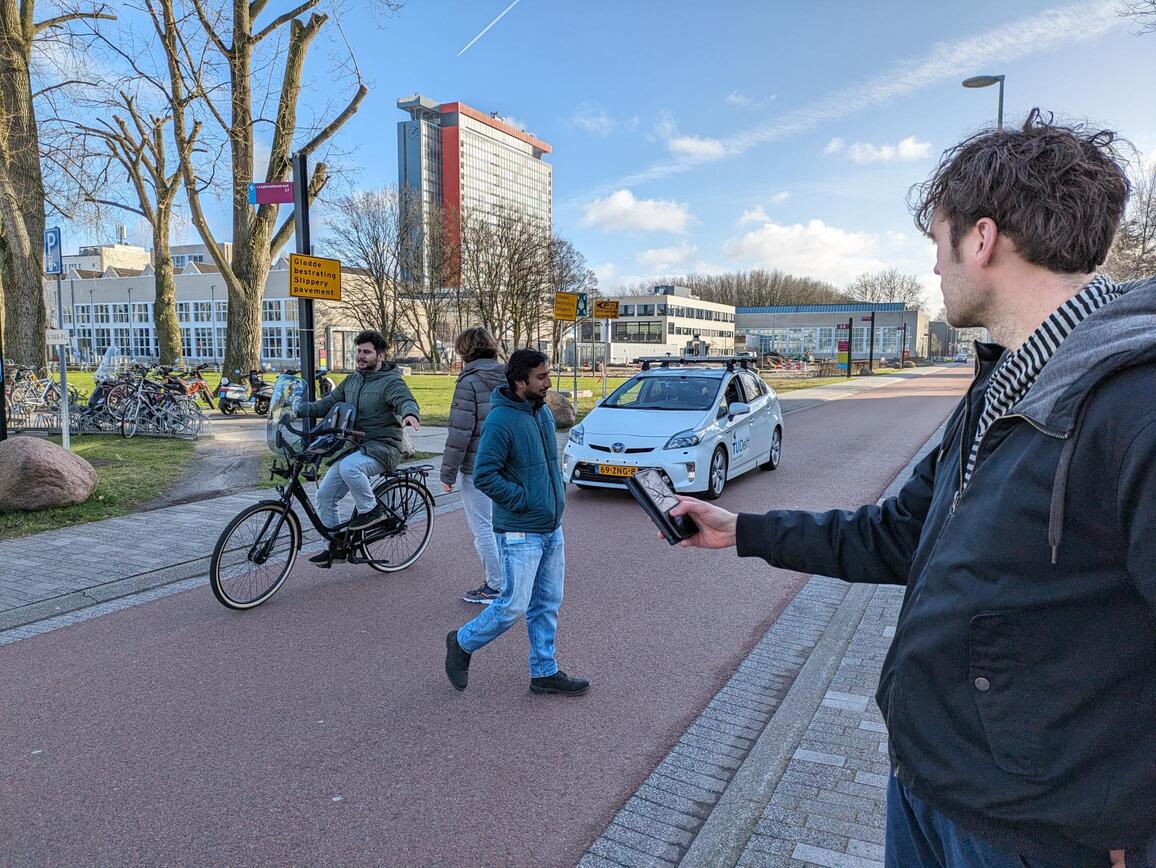}
   \caption{Safe and efficient vehicle navigation among VRUs.}
    \label{fig:demo_situation}
\end{figure}

In this paper, we present a vehicle system whose key feature is the ability to safely and efficiently navigate around Vulnerable Road Users (VRUs), such as pedestrians and cyclists (see Fig.~\ref{fig:demo_situation}). VRUs present persistent challenges for automated driving due to their high maneuverability, unpredictable behavior, and occasional disregard for traffic rules.
Our motion planner explicitly accounts for the uncertainty in predicted VRU behavior and evaluates multiple evasive maneuvers in parallel. We focus on this motion planner, without addressing broader mission-level planning. 
The Operational Design Domain (ODD) is limited to low-speed environments, such as parking lots and traffic-calmed urban areas, where multiple VRUs might interact with the self-driving vehicle.

There may be traffic situations in which autonomous operation would pose a safety risk or cause disruption to other road users. Examples include mismatches between the HD map and the real environment (e.g., due to new construction), the presence of emergency responders following an incident, or scenarios where a police officer is manually directing traffic. To handle such exceptional cases, we implement a mechanism for remote takeover, allowing a human operator to assume control of the vehicle when needed.

In the following sections, we describe related work, our prototype vehicle (an instrumented Toyota Prius), the core system modules—including environment perception, localization and mapping, motion planning, and control—and the overall system architecture.

\section{Related Work}
\label{sec:related_work}
There have been a number demonstrations on automated valet parking in academia and industry. 
For an overview, we refer the reader to~\cite{Banzhaf2017}.
Notable distinctions are in the use of a dedicated infrastructure~\cite{Min2013,Ibisch2013,Loper2013},
the use of maps that are generated offline~\cite{Jeevan2010,Stanek2010},
as well as low-cost~\cite{Furgale2013,Schwesinger2016} versus high~\cite{Urmson2009,Montemerlo2009} cost sensors. 
Many of the previous works focus on the parking aspect and assume that the environment is free of dynamically moving VRUs~\cite{Urmson2009,Montemerlo2009,Kummerle2009}.

When VRUs are present, the vehicle must take evasive action~\cite{dang2012steering}.
Trajectory optimization motion planning techniques~\cite{schwarting_safe_2018} can consider VRUs as dynamic collision avoidance constraints while optimizing performance criteria (e.g., road-following). This was demonstrated in~\cite{ferranti_safevru_2019}, which incorporated uni-modal uncertainty for VRU predictions to avoid a crossing pedestrian. Several approaches~\cite{eiras_two-stage_2022, ding_epsilon_2022} connect optimization-based planners with a high-level planner to concurrently optimize multiple driving maneuvers. They rely however on explicit lane structure and simplify the uncertainty associated with the predicted motion of other traffic participants.
Recently~\cite{de_groot_topology-driven_2024} used the underlying topology of the collision-free space to plan multiple high-level trajectories without relying on road structure, building on top of existing trajectory optimization methods. Multi-modal uncertainties of multiple VRU motion predictions were considered in~\cite{de_groot_scenario-based_2023}.

For safe vehicle remote driving, various challenges have to be addressed, such as the remote driver's situational awareness ~\cite{Zhao2023a} and embodiment.
A multitude of efforts have focused on overcoming this bottleneck, such as assisting drivers in locating vehicles~\cite{Vozar20128387}, and recognising speed~\cite{Tang201310} and distance~\cite{Tyczka20128387}. 
Recent research findings~\cite{papaioannou2023motion, zhao2023driving} have shown that vibrational feedback can increase their situational awareness and embodiment of the remote operator.

The main contributions of this system paper are:
\begin{itemize}
    \item We present a vehicle system for navigating among VRUs, which includes a novel motion planner based on~\cite{de_groot_scenario-based_2023, de_groot_topology-driven_2024} to plan several distinct probabilistic safe vehicle trajectories in parallel. Our high-level planner does not rely on an explicit lane structure, which enables its application in more unstructured urban environments.
    The underlying trajectory optimization constrains the joint probability of collision with VRUs under generic uncertainties.
    \item We furthermore integrate remote operation as a fall-back to autonomous driving. A Remote Control Tower, placed on a motion platform, provides vibrational feedback to the remote operator, in addition to video and audio, enhancing situational awareness~\cite{papaioannou2023motion, zhao2023driving}.  
\end{itemize}

\section{Vehicle System}\label{sec:system}

\begin{figure}[t]
    \centering
    \includegraphics[width=\columnwidth]{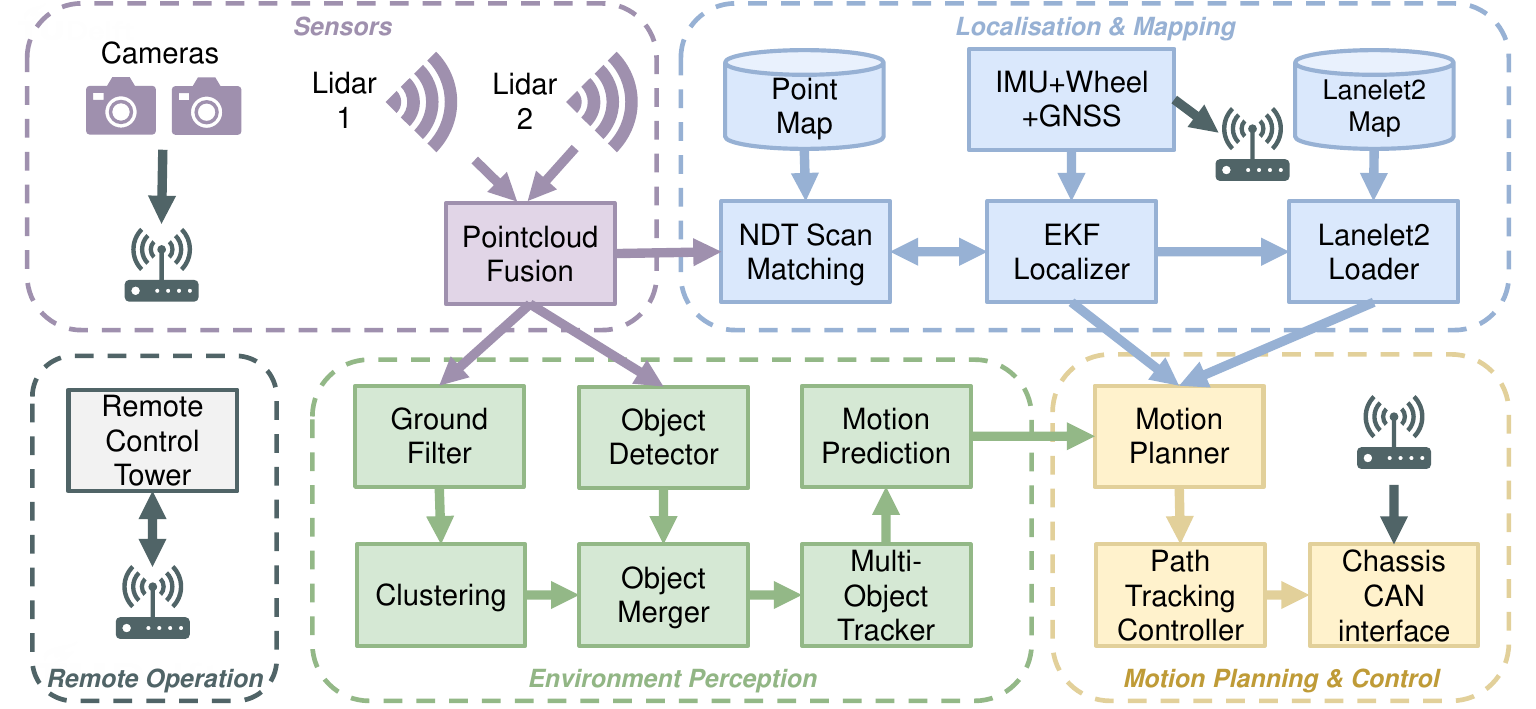}
    \caption{Overview of the demonstrator's architecture.
    The dashed-colored regions indicate the main components.
    The remote transmission symbol indicates data being exchanged between the vehicle and the external Remote Control Tower.
    }
    \label{fig:architecture}
    \vspace{-3mm}
\end{figure}

The high-level system architecture is shown in Fig.~\ref{fig:architecture}.
This section discusses the vehicle's main components: hardware and sensors (Section~\ref{subsec:prototype_vehicle}),
perception (Section~\ref{subsec:perception}),
localization and mapping (Section~\ref{subsec:localization}),
motion planning and control
(Sections~\ref{subsec:motion_planning} and \ref{subsec:control}).
Section~\ref{subsec:remote_operation} presents the remote operation.

\subsection{Prototype Vehicle}
\label{subsec:prototype_vehicle}

Our vehicle prototype is a Toyota Prius Hybrid (cf. Fig.~\ref{fig:demo_situation}). It features a custom drive-by-wire interface \emph{``Movebox''} that allows lateral and longitudinal vehicle control. On-board processing uses a PC with 
an AMD EPYC 7313 16-core processor, \qty{256}{\giga\byte} RAM, and two NVIDIA RTX A6000 GPUs.

The vehicle is equipped with two 128-beam rotating LiDARs, a Robosense Ruby and an Ouster OS1-Rev 7 both on the vehicle roof. The Robosense LiDAR, which has a larger detection range, is placed on the front at the roof, whereas the Ouster LiDAR, which has a larger vertical Field of View (\qty{45}{\degree}), is placed at an additionally elevated position, more centrally on the roof.
It also has eight Surrounding Lucid Triton Cameras and an OXTS3000 Dual antenna GNSS/INS unit.

The prototype vehicle's software stack is built on the ROS~2 Autoware\footnote{\url{https://github.com/autowarefoundation/autoware}} framework.
We use official open-source ROS~2 drivers for the LiDARs\footnote{\url{https://github.com/ouster-lidar/ouster-ros}}\footnote{\url{https://github.com/RoboSense-LiDAR/rslidar_sdk}}, cameras\footnote{\url{https://github.com/lucidvisionlabs/arena_camera_ros2}} and GNSS/INS\footnote{\url{https://github.com/OxfordTechnicalSolutions/oxts_ros2_driver}}.
For the \emph{``Movebox''} interface, we developed our own ROS~2 driver, which translates Autoware control commands (ROS~2 messages) to CAN-BCM messages and sends them to the Electronic Control Unit~(ECU) of the vehicle. It also reads the state of the car (\emph{i.e.} Control Mode, Gear, Steering angle, etc.) from the ECU and translates these signals back to high-level ROS~2 messages.
For the motion planner (see Subsection~\ref{subsec:motion_planning}) and vehicle controller (see Subsection~\ref{subsec:control})
new ROS~2 packages were developed, which will be publicly released\footnote{The planner is available at: \url{https://github.com/tud-amr/mpc_planner}}.

\subsection{Perception}
\label{subsec:perception}
We employ a modular LiDAR perception pipeline — based on Autoware~\cite{Kato2018} — to detect and track objects and predict their future trajectories.
Specifically, we perform ego-motion compensation on the LiDAR point cloud that covers the \qty{360}{\degree} surroundings of the vehicle.
To detect objects from a set of predefined classes, as well as generic objects, we use two detection components.
A deep learning-based detector detects objects from a set of predefined classes, e.g. car, bike, and pedestrian.
We use a traditional LiDAR clustering pipeline to segment generic objects that are not part of the above set of classes.
These object candidates are merged, tracked across time, and a motion prediction component predicts the future path of each tracked object.

\begin{figure}[h]
    \centering
    \includegraphics[width=0.95\columnwidth]{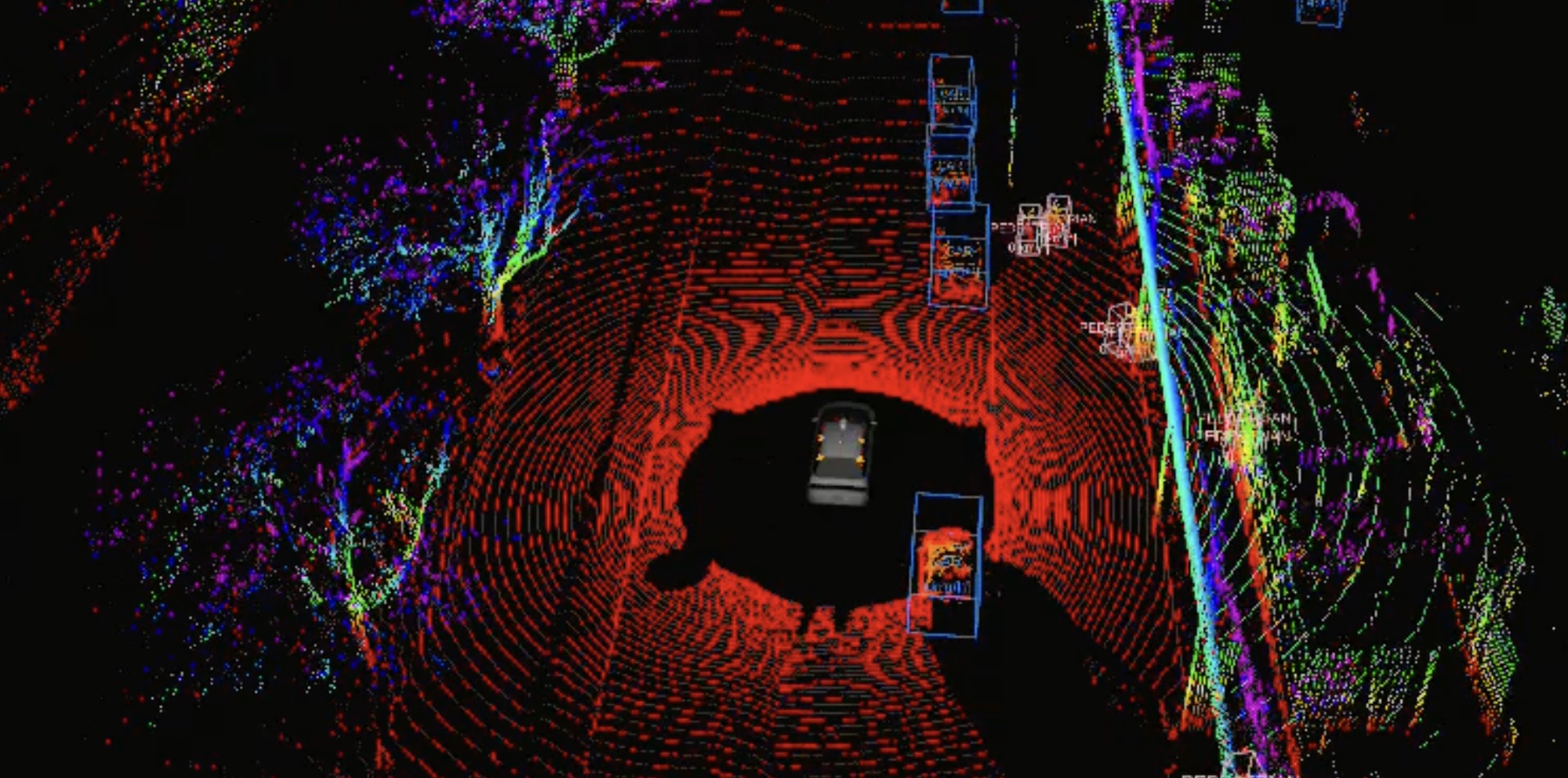}
    \caption{Visualization of the object detections. The detector detects a predefined set of object classes (e.g. car and pedestrian).}
    \label{fig:perception_centerpoint}
\end{figure}

\begin{figure}[h]
    \centering
    \includegraphics[width=0.95\columnwidth]{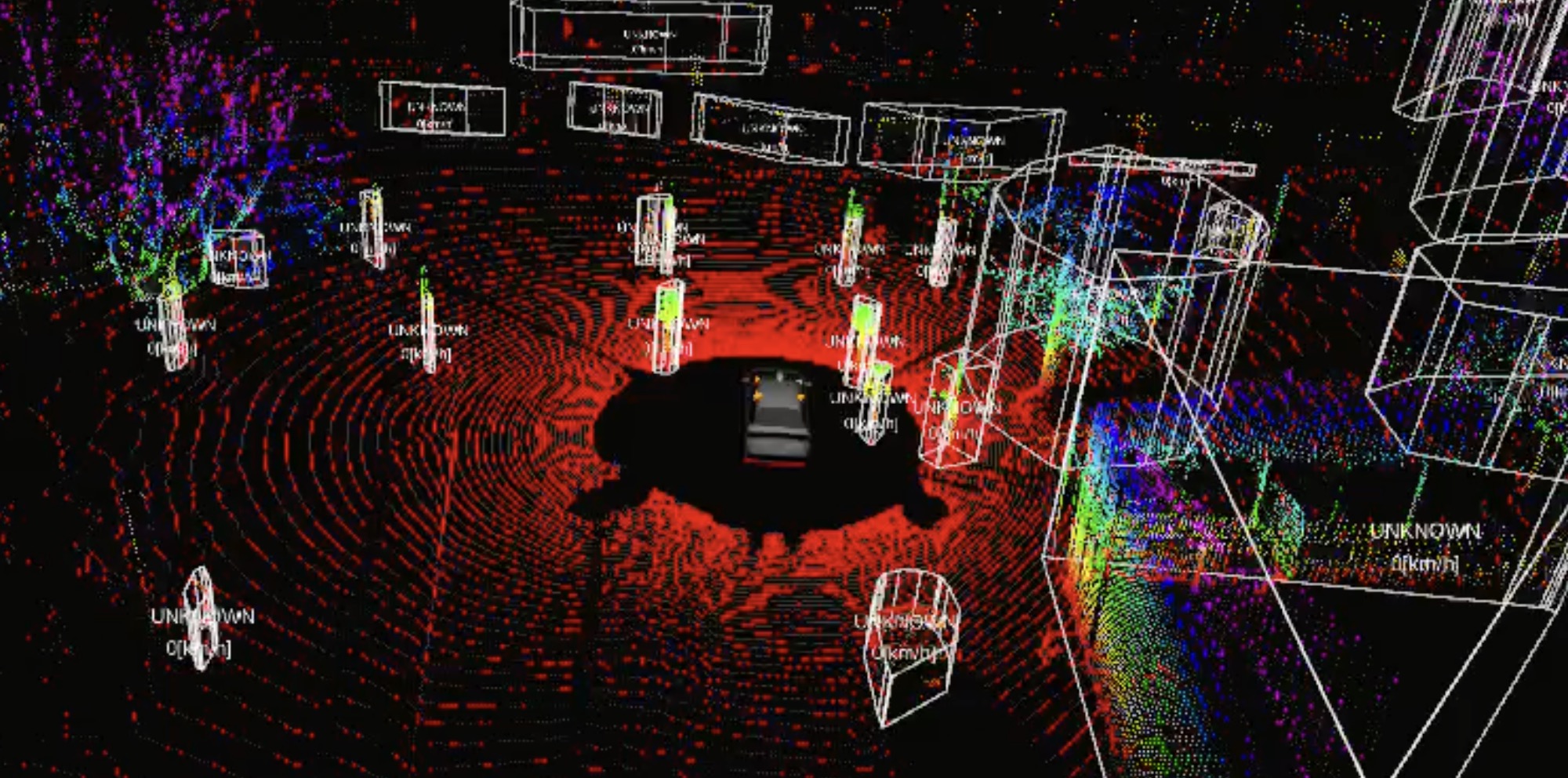}
    \caption{LiDAR clustering visualized. It segments any object protruding from the ground plane, including generic objects.}
    \label{fig:perception_clustering}
\end{figure}

\textbf{Object Detection.}
We employ the learnable 3D object detector CenterPoint~\cite{yin2021center} due to its strong performance and real-time capabilities.
CenterPoint is trained on the nuScenes~\cite{caesar2020nuscenes} dataset.
An example of the resulting detections is shown in Fig.~\ref{fig:perception_centerpoint}.
In future work, auto-labeling methods such as UNION~\cite{lentsch2024union} may be used to retrain the detector on newly collected data, eliminating the need for manual annotation.

\textbf{LiDAR clustering.}
Given a LiDAR point cloud, we remove the ground plane~\cite{shen2021fast} and group the remaining points into several class-agnostic clusters~\cite{RusuDoctoralDissertation}.
Then, we fit a bounding box for each cluster using~\cite{zhang2017efficient}.
We tune the clustering hyperparameters qualitatively on a small set of self-collected data.
Fig.~\ref{fig:perception_clustering} shows the clustering results for a scene.

\textbf{Object Merging.}
To avoid duplicate detections from the object detector and the LiDAR clustering step, we merge their outputs, which can be regarded as a minimum-cost flow problem (MCFP).
We solve the MCFP using the successive shortest path algorithm~\cite{orlin1997polynomial}. 
Matched objects are assigned labels derived from the CenterPoint detection pipeline.
Detections that do not have a corresponding match are labeled as ``unknown objects''.

\textbf{Multi-Object Tracking.}
We apply a multi-object tracker on the merged objects to smooth tracks, infer the track identities, and estimate the velocity of each object.
The data association module performs maximum score matching using muSSP~\cite{wang2019mussp} as a solver, to associate objects from neighboring frames. 
We build a separate Extended Kalman Filter (EKF) tracker for each learned object class, as well as the unknown object class.

\textbf{Motion Prediction.}
We use the EKF proposed in the previous step and propagate the uncertainty $4$ seconds into the future.
These unimodal distributions represent the future locations of all actors in the scene.
The predicted distributions are fed to the motion planner (see Subsection~\ref{subsec:motion_planning}).

\begin{figure}[t]
    \centering
    \includegraphics[width=0.41\textwidth]{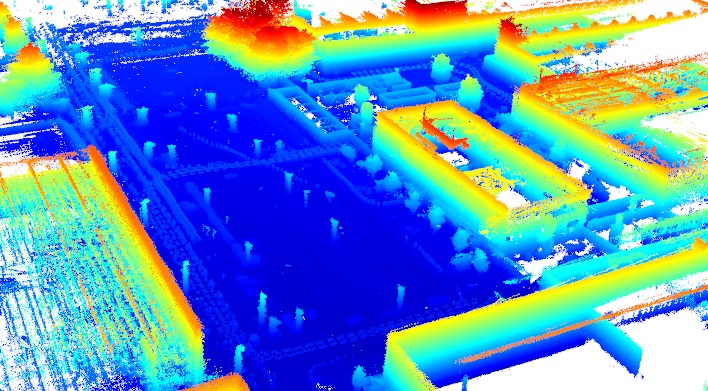}
    \caption{The point cloud map used for map-based localization in our demo.
    A point's hue indicates its height in the z-direction.
    }
    \label{fig:pcd_map}
\end{figure}

\begin{figure*}[t]
    \centering
    \includegraphics[width=\textwidth]{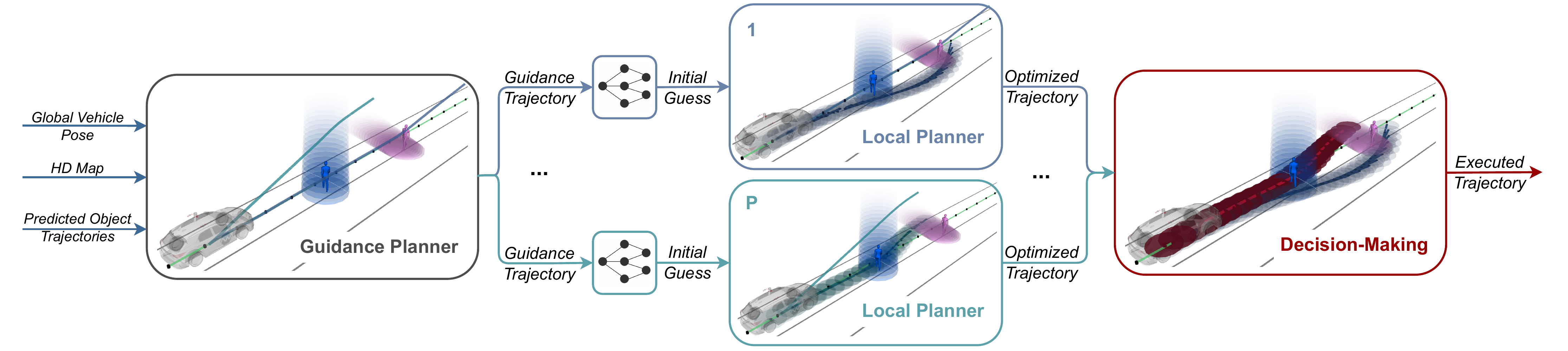}
    \caption{An overview of the motion planner (Sec.~\ref{subsec:motion_planning}). The {\color{guidance_color}\textbf{Guidance Planner}} plans several guidance trajectories that each pass the dynamic obstacles differently. Each guidance trajectory initializes a {\color{local_color}\textbf{Local Planner}} that, considering the uncertainty associated with the predicted motion of VRUs, locally optimizes the trajectory. The \textbf{Decision-Making} 
    component selects one of the optimized trajectories as output of the motion planner.}
    \label{fig:mp_diagram}
\end{figure*}

\subsection{Localization and Mapping}\label{subsec:localization}

GNSS localization alone is not always reliable when driving in areas surrounded by buildings as these obstruct the satellite reception and produce multipath reflections that yield inaccurate location estimates.
Our Autoware-based localization stack is therefore configured to also integrate the vehicle's LiDAR scans, wheel odometry and IMU to determine its
position with respect to a geo-localized 3D reference map.

\textbf{Offline Map Creation.}
The map is created offline prior to driving.
First, a 3D point cloud map of the environment is built with the open-source ROS~2 LiDAR SLAM package\footnote{\url{https://github.com/rsasaki0109/lidarslam_ros2}},
as shown in Fig.~\ref{fig:pcd_map}.
Afterwards, the 3D map is manually annotated with semantic information
using Autoware's online \textit{Vector Map Builder} tool in the Lanelet2~\cite{poggenhans2018lanelet2} format.
The annotations consist of polylines indicating the driving lanes, sidewalks, and parking areas required for planning the vehicle's trajectory.

\textbf{Online Localization.}
To produce a smooth trajectory for downstream motion planning, 
an EKF temporally integrates the relevant sensor measurements.
At each timestamp, the EKF first predicts the current vehicle state based on a predefined kinematics model and previous state estimate, and then updates the predicted state using the vehicle's measured \textit{pose} and \textit{twist} (i.e., linear and angular velocities).
To measure pose,
the vehicle's  most recent LiDAR scan is compared to the fixed 3D map
using NDT LiDAR scan matching~\cite{biber2003normal} 
and the pose estimate is adjusted iteratively to maximize the matching score.
The twist is measured using Eagleye~\cite{takanose2021eagleye}, which combines GNSS, wheel odometry, and IMU measurements.
We ensure that the GNSS has a clear reception during system start-up, and use the GNSS/INS-based pose estimate to initialize the EKF.

\subsection{Motion Planning}
\label{subsec:motion_planning}

The motion planner processes the information provided by localization and perception systems and outputs a trajectory to be tracked by the vehicle controller.
The objective is to achieve safe, comfortable, and time-efficient automated driving in a complex urban environment while engaging with VRUs such as pedestrians and cyclists.

Fig.~\ref{fig:mp_diagram} shows the structure of the motion planner that is based on Topology-driven Model Predictive Control (T-MPC)~\cite{de_groot_topology-driven_2024}. The guidance planner computes several trajectories that each pass obstacles in distinct ways. It applies a Visibility Probabilistic Roadmap (PRM)~\cite{simeon_visibility-based_2000} planner in $x, y$-position and time, filtering out trajectories that pass obstacles in the same way. Existing trajectories are tracked by propagating the planning graph over consecutive iterations. The piecewise-linear trajectories are smoothed by fitting cubic splines yielding several guidance trajectories (for more details, see~\cite{de_groot_topology-driven_2024}). In this work, we strengthen the filter in the visibility-PRM by differentiating non-passing trajectories in addition to left-right passing trajectories. We compute the difference between left, right and non-passing using winding numbers~\cite{berger_topological_2001}. The absolute value of the winding number for each agent indicates the ego-vehicle's passing progress, that we use to detect passing behaviors. The sign of the winding number indicates passing direction.

T-MPC optimizes the guidance trajectories by several local planners in parallel, ensuring that it is dynamically feasible and collision-free. A limitation of prior work~\cite{de_groot_topology-driven_2024} is that optimization of new guidance trajectories may be computationally demanding, leading to less feasible trajectories in practice. In this work, we learn a mapping from guidance trajectory to optimized trajectory. The network follows an encoder-decoder structure with dynamics incorporated in the decoder. The input of this network considers:
\begin{itemize}
    \item Discrete-time positions of one of the guidance trajectories over the time horizon,
    \item Initial ego-vehicle state,
    \item Predicted obstacle positions over the horizon each encoded with the same Long Short-Term Memory (LSTM)
    \item Reference path defined by $x$ and $y$ components, each as a sequence of cubic splines.
\end{itemize}
The decoder outputs jerk and steering rate control signals over the time horizon that are Euler integrated to retrieve trajectories. The loss of the network penalizes the deviation of state and input trajectories from the optimized ground truth obtained by running a local planner. The mapping is trained in simulation, repeatedly collecting the above inputs together with the local planner outputs over time horizon $N$.

In a simulation example, the learned model, trained with simulation data, reduced the infeasibility of guided trajectories from $26\%$ to $10\%$, greatly improving the performance of the planner. T-MPC passes the lowest-cost feasible trajectory to the vehicle control module.

In the local planners, we incorporate Gaussian uncertainty associated with the predicted motion of obstacles by using Chance Constrained MPC (CC-MPC)~\cite{zhu_chance-constrained_2019}.
CC-MPC bounds the probability of collision with each obstacle at each time step by a risk $\epsilon$. It solves the following trajectory optimization, online:
\begin{subequations}\label{eq:sh-mpc}
\begin{align}
    \min_{\b{u} \in \mathbb{U}, \b{x}\in\mathbb{X}} \hspace{1em} & \sum_{k = 0}^N J(\b{x}_k, \b{u}_k)\label{eq:optimization}\\
        \textrm{s.t.} \hspace{2em} & \b{x}_0 = \b{x}_{\textrm{init}} \label{eq:init_condition} \\
& \b{x}_{k + 1} = f(\b{x}_k, \b{u}_k), \ k = 0, \hdots, N-1 \label{eq:dynamics}\\
        &g(\b{x}_k, \b{o}^j_k) \leq 0, \: \forall k, j.\label{eq:constraints}
\end{align}
\end{subequations}

The objective $J$ includes contouring and lag costs $J_c$ and $J_l$, respectively, that track the reference path~\cite{schwarting_safe_2018} (constructed from lanelet information), a path preview cost $J_p$~\cite{chen_design_2022} to align the vehicle with the reference path at the end of the horizon, a term $||v-v_{\textrm{ref}}||_2^2$ to track a reference velocity, and penalties $||a||_2^2$ and $||\omega||_2^2$ on the commanded input and steering rate, respectively. Constraints \eqref{eq:init_condition} and \eqref{eq:dynamics} constrain the initial state and vehicle dynamics. Probabilistic collision avoidance with obstacles is specified through \eqref{eq:constraints}.

We solve the optimization problem in \eqref{eq:optimization}-\eqref{eq:constraints} using the Sequential Quadratic Programming (SQP) solver provided by FORCES Pro\cite{domahidi_forces_2014}, replanning the vehicle trajectory at \qty{10}{\hertz}.
Each planning cycle operates over a \qty{7}{\second} time horizon, discretized into $N = 35$ steps with \qty{0.2}{\second} intervals.
The vehicle dynamics follow a kinematic bicycle model.

\subsection{Control}
\label{subsec:control}
The vehicle controller computes acceleration and steering commands from the motion planner’s optimised path to ensure safe trajectory tracking. A Model Predictive Contouring Control (MPCC) is proposed to consider the coupled longitudinal and lateral dynamics \cite{bertipaglia2023model}. Its prediction model relies on a nonlinear single-track vehicle with a Fiala tyre model, parametrised with experimental data to capture also the nonlinear behaviour. The MPCC computes optimal steering and longitudinal forces to minimise tracking and velocity errors while enforcing stability. The cost function is defined as follows:

\begin{equation}
    \begin{split}
        J_{pf} = \sum_{i=1}^{N} \biggl( &q_{e_\textrm{Con}} e_{\textrm{Con}, i}^2 + q_{e_\textrm{Lag}} e_{\textrm{Lag}, i}^2 + q_{e_\textrm{Vel}} \left(e_\textrm{Vel}\right)^2 \\
        &+q_{\dot{\delta}} \dot{\delta_i}^2 + q_{\dot{F_x}} \dot{F_{x_i}}^2 \biggr), 
    \end{split}
    \label{eq:cost}
\end{equation}
where $N$ is the length of the prediction horizon, the contouring error $e_\textrm{Con}$, the lag error $e_\textrm{Lag}$, the difference between the vehicle's velocity and the desired one $e_\textrm{Vel}$, the steering rate $\dot{\delta}$, the rate of longitudinal force $\dot{F}_{x}$, and parameters $q_*$ are the weights of the respective quadratic errors.
The weights are fine-tuned to optimize controller performance by minimizing longitudinal velocity error, and trajectory offset \cite{Bertipaglia2022Two}, 
The cost function is constrained based on actuator limits, and to take into account diverse weather conditions, the total tire force at each axle is constrained within the tire friction circle to ensure vehicle stability.

The steering angle is sent directly to the logic of the column-assist electric motor (function of lane-keeping assist) as a reference via CAN. For acceleration control, the low-level controller consists of a feedforward component considering powertrain steady-state characteristics and braking dynamics and a feedback component to compensate for the difference between the commanded and actual accelerations.

\subsection{Remote Operation}
\label{subsec:remote_operation}
The remote operation takes over vehicle control in edge cases, where the vehicle cannot correctly match the perceived information with the known datasets (e.g., if the epistemic uncertainty of the perception component becomes too large~\cite{itkina_interpretable_2023}).

\begin{figure}[t]
   \centering
    \includegraphics[width=\linewidth]{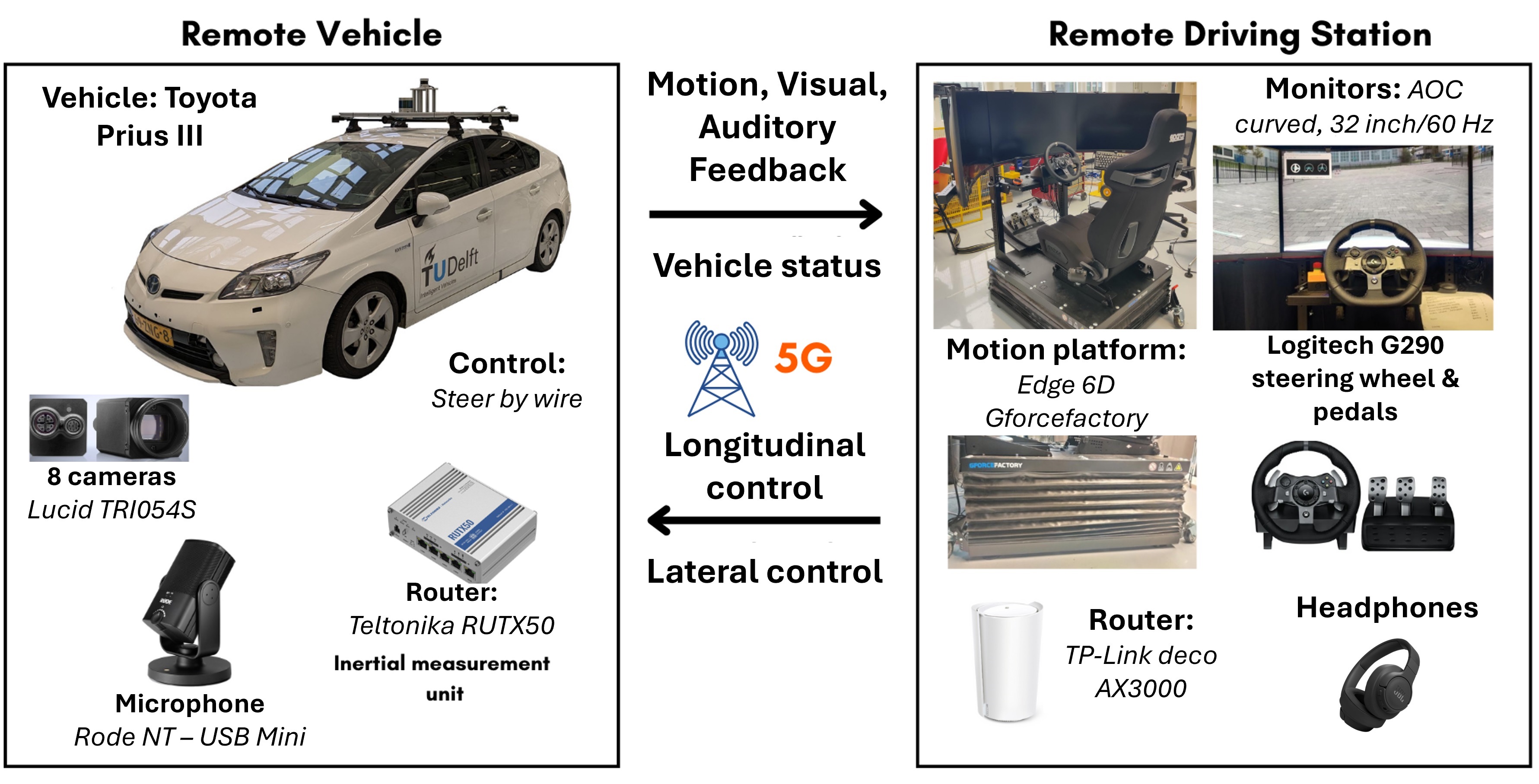}
    \caption{Remote Control Tower (RCT) with a Gforcefactory Edge 6D motion simulator equipped with three monitors.
    }
    \label{fig:RCT_Setup}
\end{figure}

To that end, a Remote Control Tower (RCT) is connected with the prototype vehicle, and receives information provided by the experimental vehicle (Fig. \ref{fig:architecture}).
The RCT is comprised of several components: a motion
platform, three high-resolution monitors (curved 32 inch, 60Hz), a steering wheel, pedals, and a racing seat. 
Each of these elements contributes to creating an immersive and realistic driving experience for the remote operators.
The motion platform is a Gforcefactory EDGE 6D motion simulator\footnote{\url{https://www.gforcefactory.com/edge-6d}} (Fig.~\ref{fig:RCT_Setup}), running with a UDP protocol at \qty{200}{\hertz}.
Three low-latency AOC monitors (\qty{1}{\milli\second}) are used to provide video feedback.
For controlling the steering and throttle of the vehicle, we have used the Logitech G920 steering wheel and pedals\footnote{\url{https://www.logitechg.com/en-us/products/driving/driving-force-racing-wheel.html}}.
For a more realistic steering feel, force feedback is implemented in the form of a spring and auto centering.
The steering wheel has a small motor which can generate different types of feedback. This motor can be activated using $ff_{effect}$ from the force feedback framework for Linux \footnote{\url{https://www.kernel.org/doc/Documentation/input/ff.txt}}. 

Within the RCT, auditory feedback is provided to the operator via noise-canceling headphones. 
The headphones receive the sound from the car captured by a RODE NT-USB Mini microphone \footnote{\url{https://rode.com/de/microphones/usb/nt-usb-mini}} which is located directly next to the driver.
In the experimental vehicle, multiple cameras are used to collect high-resolution real-world video.
To further enhance the situational awareness of the remote driver, the motion platform also provides the operator with vibrational feedback from the experimental vehicle~\cite{papaioannou2023motion, zhao2023driving}.

The RCT and the vehicle maintain connectivity through a 5G network, using Hyperpath for low-latency communication \footnote{Hyperpath - \url{https://www.hyperpath.ie/}}. 
Hyperpath creates a multi-connectivity peer-to-peer mesh VPN (Virtual Private Network) between the vehicle and the RCT. 
The protocol used is the User Datagram Protocol (UDP). 
For real-time streaming of camera footage, WebRTC is used \cite{robotwebtools-no-date}.
In this way, all required signals are transmitted i.e., actuation inputs (steering wheel and throttle input) and feedback (video, audio, and vibrational).

\begin{figure*}[t]
    \centering
    \begin{subfigure}[t]{0.32\textwidth}
        \includegraphics[width=\textwidth]{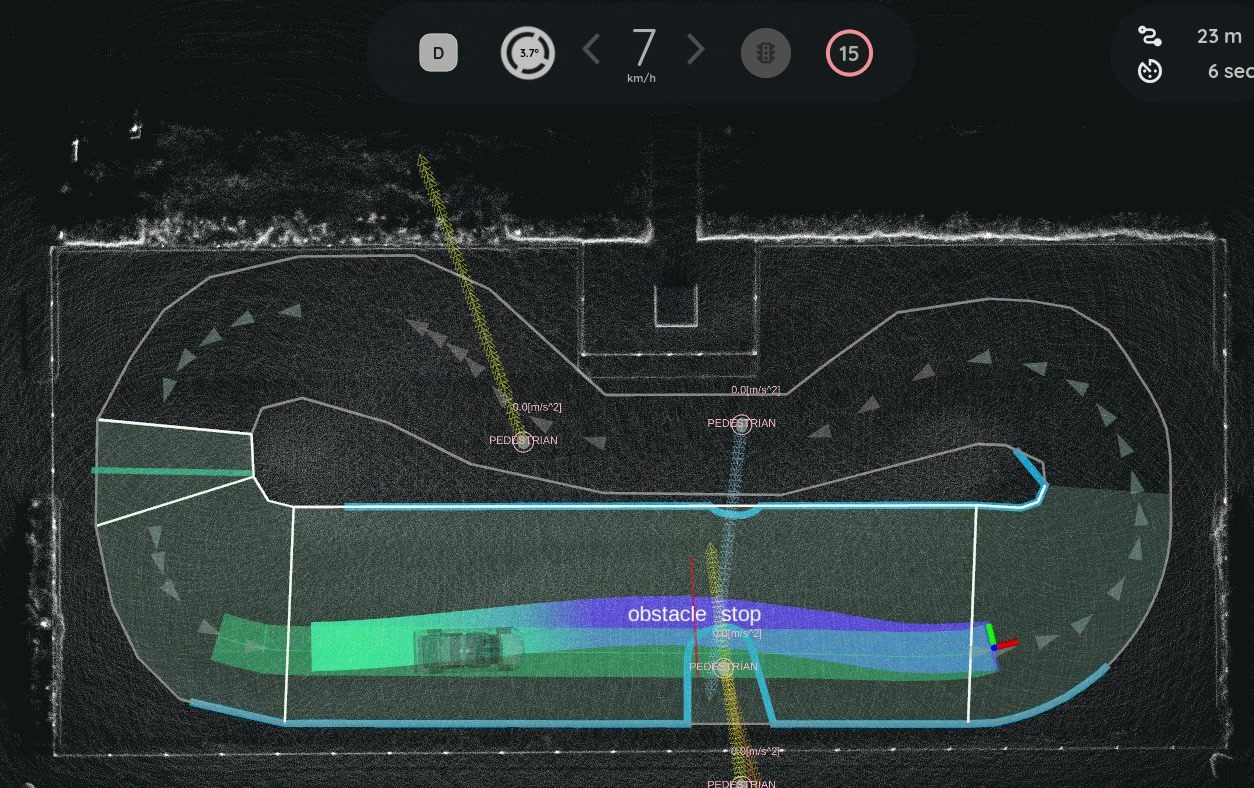}
        \caption{The Autoware planner brakes for the obstacles.}
     \end{subfigure}
     \hfill
     \begin{subfigure}[t]{0.32\textwidth}
        \includegraphics[width=\textwidth]{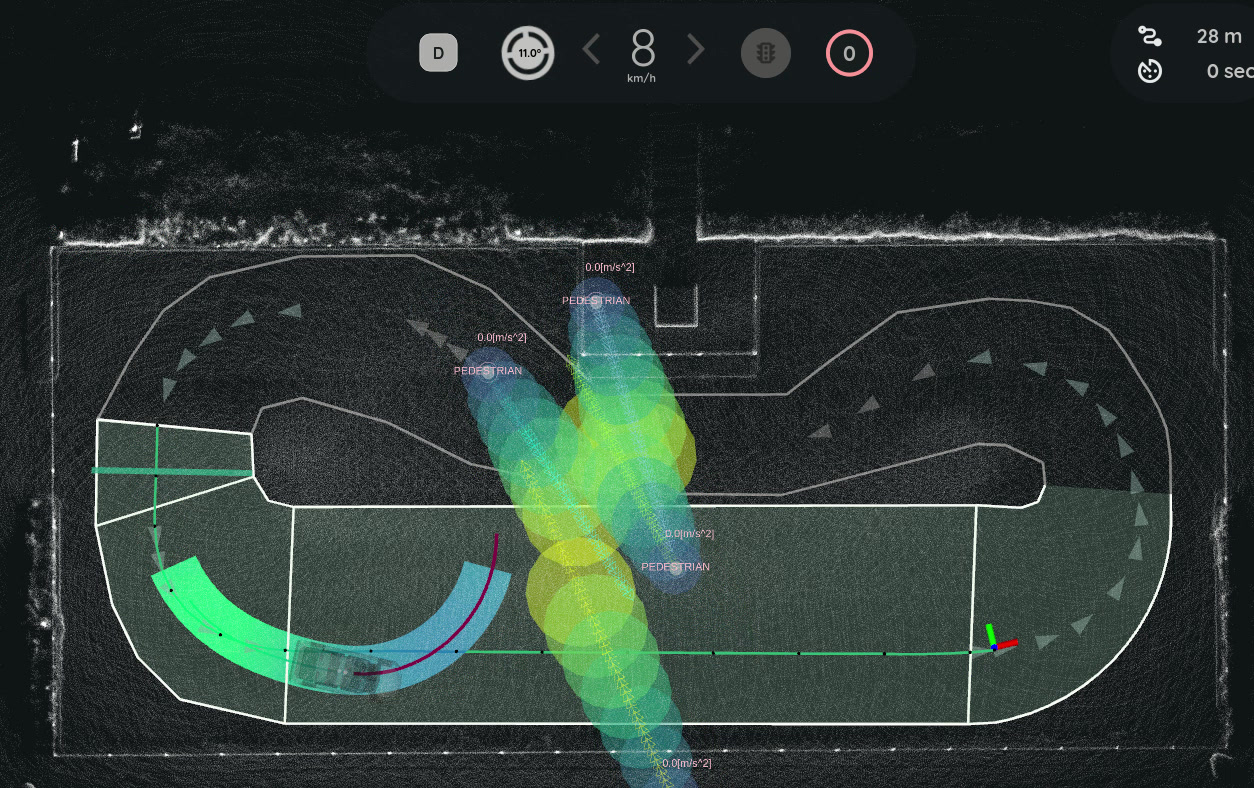}
        \caption{LMPCC finds a poor local optimal trajectory that brakes for the obstacles.}
     \end{subfigure}
    \hfill
     \begin{subfigure}[t]{0.32\textwidth}
        \includegraphics[width=\textwidth]{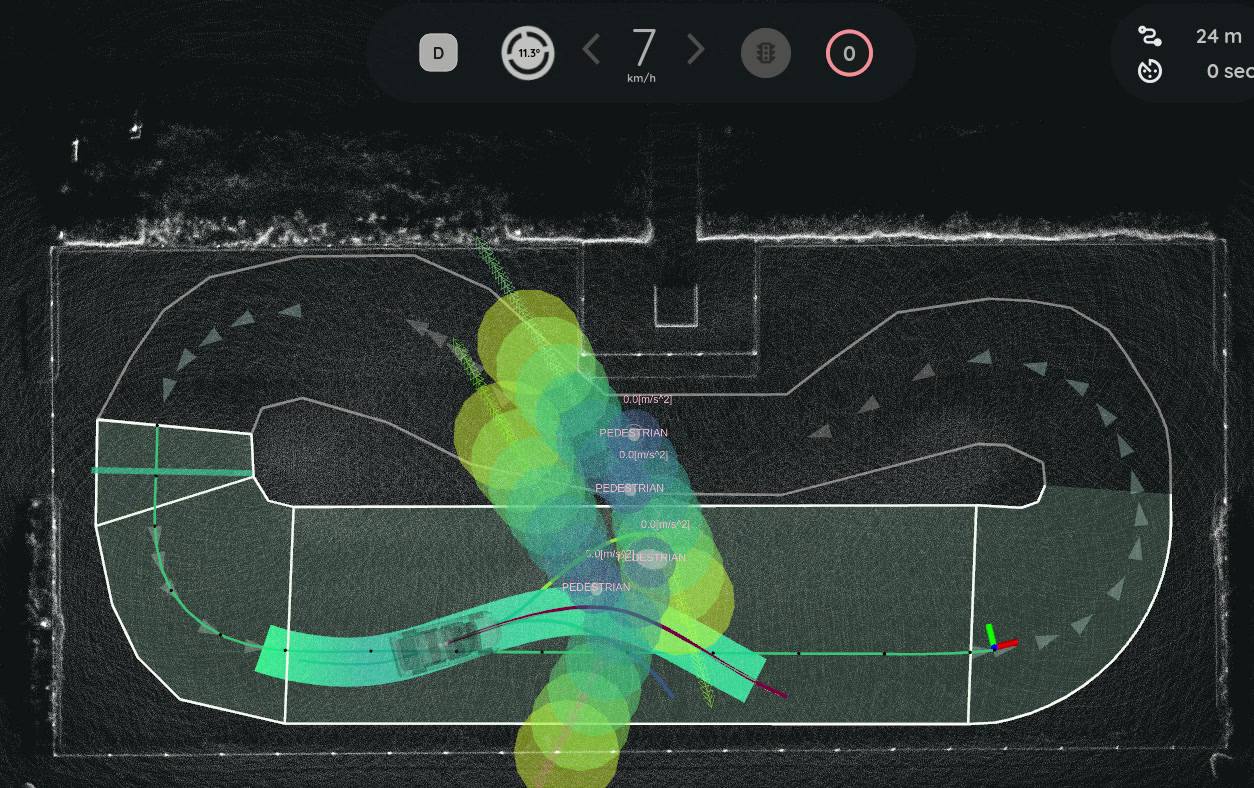}
        \caption{T-MPC\texttt{++} steers to evade the obstacles efficiently.}
     \end{subfigure}
    \caption{Screenshot of the simulations where the vehicle is following annotated lanelets (green solid line) in the vicinity of pedestrians (marked pink). Pedestrian predictions (sequence of circles) are visualized including level sets of the associated uncertainty in (b, c). Vehicle planned trajectory visualized in blue (slow) to green (fast). Alternative topological trajectories in (c) are shown with solid colored lines.}
    \label{fig:screenshots}
\end{figure*}

\begin{figure}[t]
    \centering
\includegraphics[width=0.95\columnwidth]{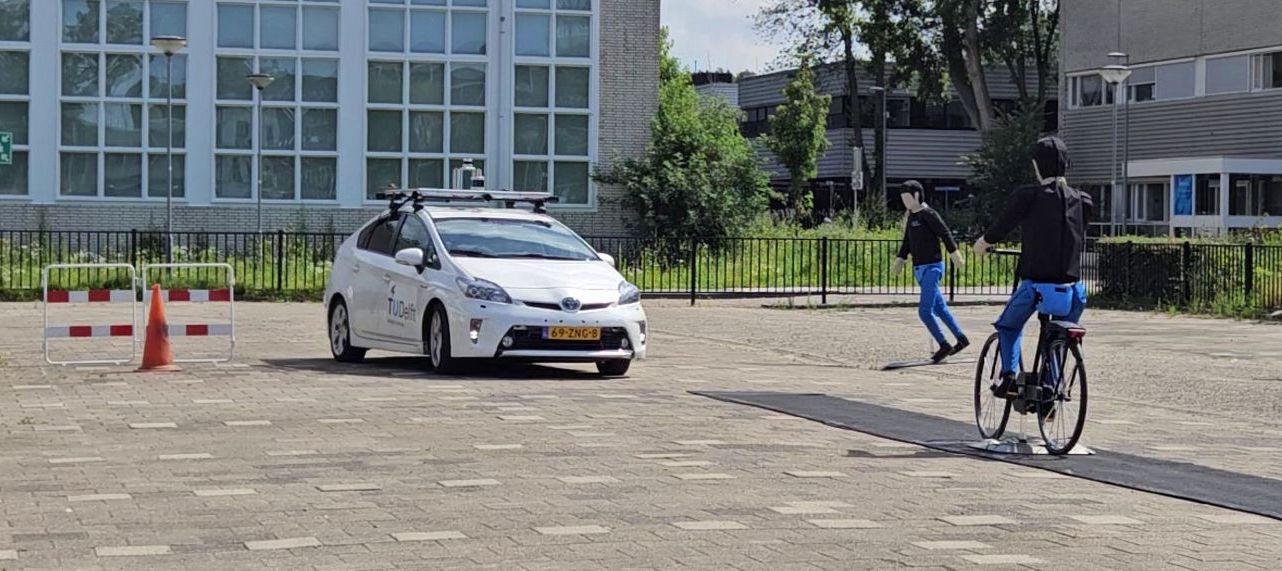}
    \caption{A snapshot of the vehicle tests.} \label{fig:rss2024demo}
\end{figure}

\begin{figure}[h]
   \centering
    \includegraphics[width=\linewidth]{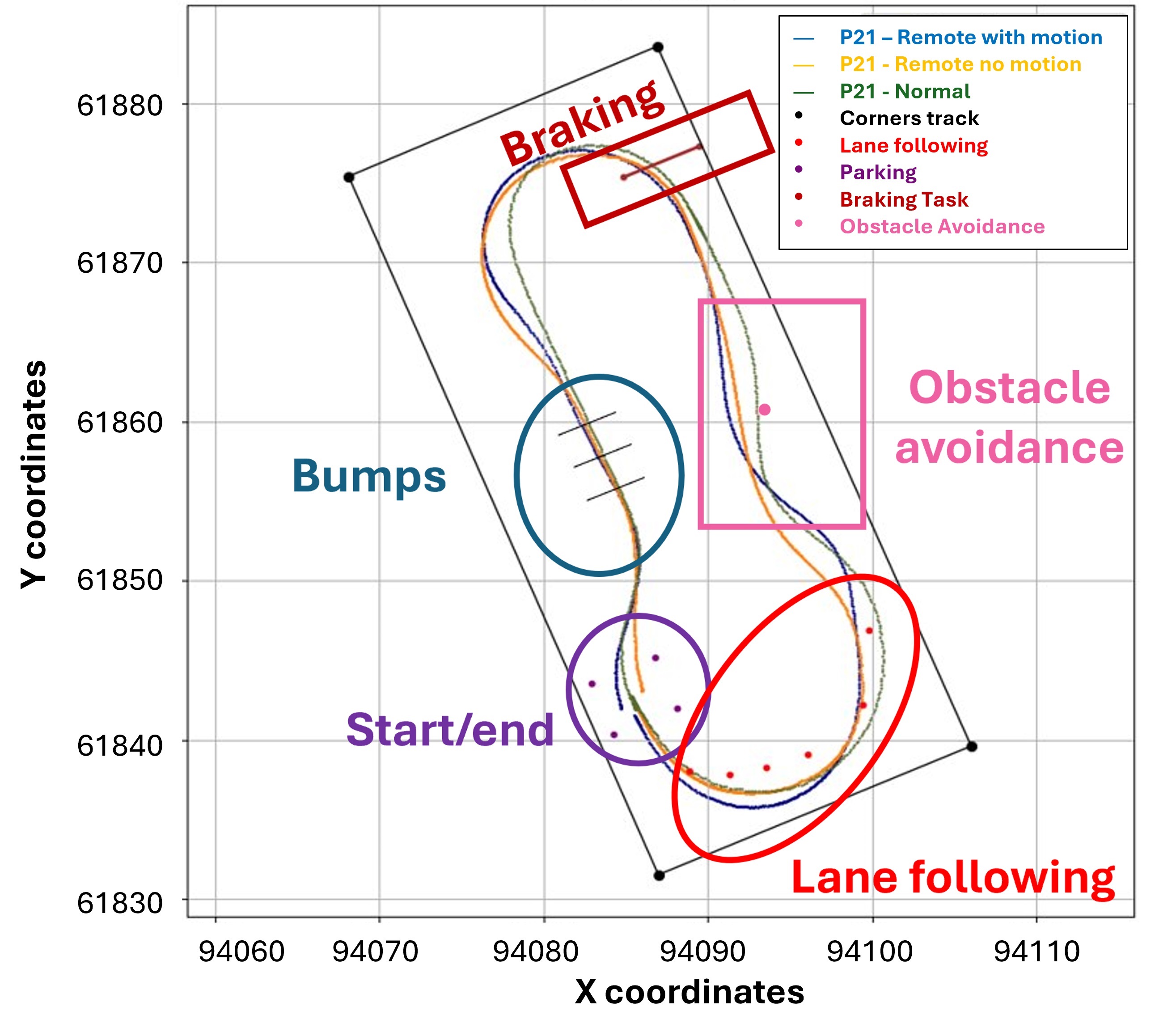}
    \caption{GNSS data of one participant, lap 4 of all conditions, including locations of tasks.}
    \label{fig:gnssdata}
\end{figure}

\section{Experiments}
\label{sec:Demonstration}
To evaluate overall driving performance, we compare our proposed motion planner in simulation to three baselines:

\begin{itemize}
    \item \textbf{Autoware Planner}\footnote{see \url{https://autowarefoundation.github.io/autoware-documentation/main/design/autoware-architecture/planning/}}~\cite{Kato2018}: The base planner included in Autoware. Its default behavior brakes for any nearby obstacle and is not competitive with other methods. To make the planner more competitive, we enabled dynamic collision avoidance and tuned the planner (lower safety margins, prefer overtaking over braking) to avoid the pedestrians as much as possible in our simulation environment.
    \item \textbf{LMPCC}~\cite{ferranti_safevru_2019}: A non parallelized MPC implementation.
    \item \textbf{T-MPC\texttt{++} (w/o fallback)}~\cite{de_groot_topology-driven_2024}: Parallelized MPC without fallback trajectory classes and learned initial guesses.
\end{itemize}

\noindent All optimization-based methods have identical objectives and collision avoidance constraints. We compare these methods to our proposed method in scenarios with $0, 2$ and $4$ pedestrians to study how well these methods cope with increasingly crowded dynamic environments. Quantitative results are shown in Table~\ref{tab:baseline_comparison} and screenshots are shown in Fig.~\ref{fig:screenshots}.

\begin{table}[b]
\small\sf\centering
\caption{Comparison of navigation performance on a scenario with a varying number of randomized crossing pedestrians over $25$ experiments. Reported is the task duration, number of collisions, time-outs (vehicle did not reach the goal in time) and average velocity.}
\resizebox{0.49\textwidth}{!}{%
\begin{tabular}{|l|l|c|c|c|c|}
\hline\textbf{\#} & \textbf{Method} & \textbf{Dur. [s]} & \textbf{Collisions} & \textbf{Time-outs} & \textbf{Avg. Vel. [m/s]} \\\hline
\multirow{2}{*}{0} & Autoware & 19.5 (0.1) & \textbf{0} & \textbf{0} & \textbf{1.87} \\
&T-MPC\texttt{++} & \textbf{19.4} (0.3) & \textbf{0} & \textbf{0} & 1.86 \\\hline
\multirow{4}{*}{2} & Autoware & 24.0 (4.2) & 1 & \textbf{0} & 1.54 \\
&LMPCC & 22.4 (3.3) & \textbf{0} & \textbf{0} & 1.65 \\
&T-MPC\texttt{++} (w/o fallback) & \textbf{21.5} (1.6) & \textbf{0} & \textbf{0} & \textbf{1.70} \\
&T-MPC\texttt{++} & 21.6 (1.8) & \textbf{0} & \textbf{0} & \textbf{1.70} \\\hline
\multirow{4}{*}{4} & Autoware & 27.9 (9.7) & 2 & 3 & 1.22 \\
&LMPCC & 26.7 (6.6) & 1 & 2 & 1.43 \\
&T-MPC\texttt{++} (w/o fallback) & 24.0 (3.3) & \textbf{0} & \textbf{0} & 1.57 \\
&T-MPC\texttt{++} & \textbf{23.5} (3.8) & \textbf{0} & \textbf{0} & \textbf{1.61} \\\hline
\end{tabular}}
\label{tab:baseline_comparison}
\end{table}

The Autoware baseline is the slowest planner and often collides.
LMPCC (i.e., single trajectory MPC) collides less and drives faster on average. It is outperformed however by both versions of T-MPC++, in particular by comparing the task durations with an increasing number of obstacles. With $4$ obstacles, on average LMPCC takes $7.3$s longer, while T-MPC++ takes only $4.1$s longer (i.e., T-MPC++ is more than $50$\% faster in dealing with $4$ obstacles in this environment). T-MPC++ therefore passes the obstacles more efficiently.

We deployed our motion planner on our prototype vehicle to navigate around a static obstacles and a moving dummy cyclist on our (closed) test track. The vehicle's task is to follow encoded lanelets, while avoiding the obstacles. Fig. \ref{fig:rss2024demo} shows snapshot from one of the many runs. Our vehicle was able to safely and efficiently avoid the obstacles.

To evaluate the remote driving performance, we considered three conditions.
Participants control the vehicle:
(a) from the vehicle driver's position; 
(b) from the RCT receiving motion feedback
(c) from the RCT without any motion feedback, relying solely on visual and auditory feedback.
The participants drove five laps around the test track for each condition.
The paths driven by one participant are illustrated in Fig.~\ref{fig:gnssdata}.
The values measured for the velocity, steering position, longitudinal and lateral acceleration were significantly higher (p $<$ 0.004) in both conditions of remote driving (motion/no motion feedback) compared with in-vehicle
driving. 
No significant difference was identified between the remote driving conditions (motion/no motion feedback) for any of the metrics stated above.
Regarding the driving experience, all items included in our questionnaire were significantly higher (p$<$0.03) for in-vehicle driving compared to both remote driving conditions. 
The findings are further detailed in \cite{schippers2024motion}, and suggest that motion feedback, in its current form, may not be necessary for low-velocity scenarios. 
Thus a simplified RCT without motion feedback could suffice for safe driving in these scenarios. Additionally, the results show that there is still a large gap between remote driving and in-vehicle driving both in the overall driving performance and experience, but also in the successful completion of the specific tasks.

\section{Conclusions}\label{sec:future_work}
We presented a vehicle system that is able to navigate
safely and efficiently around Vulnerable Road Users (VRUs). One distinguishing aspect is a novel motion planner, based on Topology-driven Model Predictive Control (T-MPC). The guidance planner computes several trajectories in
parallel that each pass obstacles in distinct way, or are non-
passing. The underlying trajectory optimization constrains the
joint probability of collision with VRUs under generic uncertainties. To account for ``edge'' cases that the vehicle might not be able to handle autonomously, the
system included remote operation.

In simulation, we showed that our motion planner
outperforms three baselines in terms of safety and efficiency.
Furthermore, we described tests with the prototype vehicle on a
test track in self-driving mode and with remote operation. Future work includes more sophisticated trajectory prediction, a complete mission planner, improved situational awareness of the remote operator at higher speeds, and a self-assessment method for switching from autonomous to remote driving.

\section*{Acknowledgments}
This research has been conducted as part of the EVENTS project, which is funded by the European Union, under grant agreement No 101069614. Views and opinions expressed are, however, those of the author(s) only and do not necessarily reflect those of the European Union or European Commission. Neither the European Union nor the granting authority can be held responsible for them.
In addition, we thank T. de Boer and H. Harmankaya for their support on remote operation.


\balance
\bibliographystyle{unsrt}
\bibliography{references}

\end{document}